\documentclass[lettersize,journal]{IEEEtran}
\usepackage{amsmath,amsfonts}
\usepackage{algorithmic}
\usepackage{algorithm}
\usepackage{array}
\usepackage[caption=false,font=normalsize,labelfont=sf,textfont=sf]{subfig}
\usepackage{textcomp}
\usepackage{stfloats}
\usepackage{url}
\usepackage{verbatim}
\usepackage{graphicx}
\usepackage{cite}
\usepackage[colorlinks,            linkcolor=black,        anchorcolor=black,     citecolor=black            ]{hyperref}

\newtheorem{theorem}{Theorem}
\newtheorem{lemma}{Lemma}

\usepackage{xcolor}
\usepackage{array}

\begin{document}

\title{Robust Distributed Learning Against Both Distributional Shifts and Byzantine Attacks}

\author{Guanqiang Zhou, Ping Xu, Yue Wang, and Zhi Tian

\thanks{This work was partly supported by the US NSF grant CIF-1939553.}

\thanks{All the authors are with the Department of Electrical and Computer Engineering, George Mason University, Fairfax, VA 22030, USA (email: gzhou4@gmu.edu; pxu3@gmu.edu; ywang56@gmu.edu; ztian1@gmu.edu).}
}


\maketitle

\begin{abstract}
In distributed learning systems, robustness issues may arise from two sources. On one hand, due to distributional shifts between training data and test data, the trained model could exhibit poor out-of-sample performance. On the other hand, a portion of working nodes might be subject to byzantine attacks which could invalidate the learning result. Existing works mostly deal with these two issues separately. In this paper, we propose a new algorithm that equips distributed learning with robustness measures against both distributional shifts and byzantine attacks. Our algorithm is built on recent advances in distributionally robust optimization as well as norm-based screening (NBS), a robust aggregation scheme against byzantine attacks. We provide convergence proofs in three cases of the learning model being nonconvex, convex, and strongly convex for the proposed algorithm, shedding light on its convergence behaviors and endurability against byzantine attacks. In particular, we deduce that any algorithm employing NBS (including ours) cannot converge when the percentage of byzantine nodes is $\frac{1}{3}$ or higher, instead of $\frac{1}{2}$, which is the common belief in current literature. The experimental results demonstrate the effectiveness of our algorithm against both robustness issues. To the best of our knowledge, this is the first work to address distributional shifts and byzantine attacks simultaneously.
\end{abstract}


\section{Introduction}\label{sec1}
\IEEEPARstart{D}{istributed} learning plays an important role in solving large-scale machine learning problems. It refers to the paradigm where a number of working nodes (workers) carry out the overall task of training a model in parallel, coordinated by a central node (server). As the deployment of machine learning becomes prevalent in modern safety-critical fields (such as autonomous driving \cite{driving} and medical diagnosis \cite{medical}) where the cost of model failure is extremely high, it is crucial to equip the learning systems with some robust features, such that the risk of model failure is minimized.

In distributed learning, there are two major robustness issues that may pose a threat to model safety. The first issue is distributional shifts, which exposes the vulnerability of empirical risk minimization (ERM), the de facto training paradigm in machine learning. In ERM, the model is trained to minimize the training loss and then is applied to unseen data, or test data, on the key assumption that training data and test data are drawn from the same distribution. However, this assumption rarely holds in a practical scenario due to selection biases in training data \cite{Liu}, non-stationarity in the environment \cite{MIT}, or even adversarial perturbations \cite{Madry}, leaving the ERM-trained models susceptible to drastically degraded performance under some minor level of distributional shifts. This issue applies to different training settings, distributed and centralized alike.

The second issue is byzantine attacks. In a typical distributed training iteration, each worker is supposed to send its honest and accurate local update to the server, which uses the average of these local updates to refine the model. However, due to a myriad of system glitches such as data corruption, computational error, and transmission interference, a portion of workers could send unwarranted updates to the server, thus polluting the refined model \cite{Yudong}. Even worse, an adversary might intentionally insert malicious workers into the system to attack the model. Due to the difficulty in modeling each type of system error separately, as well as the concern of malicious workers, researchers in this field often model them uniformly as byzantine attacks \cite{Lamport}, where a malfunctional/byzantine worker can send arbitrary messages to the server. It is well known that even a single byzantine worker can totally invalidate the learning result and cause model failure \cite{Blanchard}.

To combat distributional shifts, the conventional approach is robust optimization where the hypothetical data shifts are restricted to be within a deterministic uncertainty set \cite{Ben-Tal,Bertsimas,Goodfellow,Madry}, and the goal is to find the optimal model for the worst-case set of data. However, these works are found to be intractable except for specially structured losses \cite{Sinha} and they tend to promote over-conservative solutions \cite{Soroosh}. Distributionally robust optimization (DRO), on the other hand, treats the data uncertainty in a probabilistic way and has been the more favored approach to dealing with distributional shifts, due to its appealing theoretical guarantees \cite{Esfahani}, computational tractability when assembled with certain metrics, and extraordinary empirical performance \cite{Boston}.

To cope with byzantine attacks, there are generally two distinct approaches. The first approach assigns each worker redundant data, and relies on this redundancy to eliminate the effect of erroneous updates \cite{DRACO,DETOX,Data}. However, this approach is known to be computationally intensive \cite{Baruch}. Also, it is incompatible with federated learning where data cannot be replicated and reassigned for user privacy. The second approach is based on robust aggregation where the averaging (of local updates) step is replaced with a robust aggregation measure, such as Krum \cite{Blanchard}, geometric median \cite{Yudong}, coordinate-wise median \cite{Yin}, iterative filtering \cite{Su}, signSGD \cite{Bernstein}, norm-based screening \cite{ITA,Ghosh,ISIT,newton}, etc. Due to its variety of aggregation rules and applicability to federated learning, robust aggregation is the mainstream approach to mitigating byzantine attacks and is also the focus of this paper.

Despite extensive efforts to address distributional shifts and byzantine attacks separately, we observe that there has not yet been any work that claims to resolve both issues simultaneously, which may encounter two hurdles. The first hurdle is that the issue of distributional shifts has been mostly considered in the centralized setting where a single machine has access to all the data. Consequently, the established approaches often lead to solving some form of convex programs, such as linear programs \cite{Soroosh}, semi-definite programs \cite{Xu}, and second-order cone programs \cite{Mehrotra}, which are not directly solvable in a distributed network where data scatter across multiple local devices. The second hurdle is that byzantine-robust approaches only have limited success in providing theoretical convergence guarantees \cite{DETOX}, since they often require strong assumptions (such as sub-exponential \cite{Yin} and sub-Gaussian \cite{Su}) on the distribution of local gradients. These assumptions become even harder to justify when training a distributionally robust model as opposed to ERM.

\textbf{Our work.}
In this paper, we aim to fill this gap by proposing a robust distributed learning algorithm that is resilient to both distributional shifts and byzantine attacks. To address the aforementioned first obstacle, we utilize a recent work on Wasserstein distributionally robust optimization \cite{Sinha}, which leads to a reformulation that can be solved in a distributed fashion (see Section \ref{sec2-2}). To bypass the second obstacle, we implement norm-based screening (NBS), a simple robust aggregation scheme. We formulate a robust property of NBS which enables us to avoid making unjustified assumptions on local gradients while providing convergence guarantees (see Section \ref{sec3}). From another perspective, these two adopted techniques equip our algorithm with robust features against distributional shifts and byzantine attacks respectively. We further derive theoretical convergence guarantees of the proposed algorithm for nonconvex, convex, and strongly convex learning problems respectively. The theoretical results offer valuable insights into the convergence behaviors of our algorithm (see Section \ref{sec6-2}), the considerations in selecting certain parameters effectively (see Section \ref{sec6-4}), and the breakpoint of NBS (see Section \ref{sec6-3}). In particular, we point out the common misconception that the breakpoint of NBS is $\frac{1}{2}$ (of workers being abnormal) and correct it as $\frac{1}{3}$. We empirically verify our algorithm's effectiveness against both distributional shifts and byzantine attacks on the Spambase dataset \cite{spambase}, and our algorithm's outstanding performance is shown to be not sensitive to the selection of hyper-parameters.
	
	Our main contributions are summarized as follows:
	\begin{enumerate}
		\item We propose a distributed learning algorithm with robust features against both distributional shifts and byzantine attacks, the very first of its kind.
		\item We provide convergence proofs for our algorithm for nonconvex, convex, and strongly convex learning problems respectively, giving insights into our algorithm's convergence behaviors, endurability against byzantine attacks, and parameter-selection strategies.	
		\item For the first time, we debunk the widely held misconception that the breakpoint of NBS is $\frac{1}{2}$, and we deduce that it should have been $\frac{1}{3}$.
	\end{enumerate}

\textbf{Notations.}
Throughout, the norm notation $\|\cdot\|$ refers to the $L_2$ norm if not otherwise specified.

\section{Related Work}
\subsection{Distributionally Robust Optimization (DRO)}
DRO is a principled methodology of handling distributional shifts. Its objective is to find a model $\theta$ that minimizes the worst-case expected loss $\sup_{Q\in\Omega}\mathbb{E}_{x\sim Q}f(\theta;x)$ over an ambiguity set $\Omega$ which encompasses a cluster of data distributions. In practice, $\Omega$ is constructed based on the information of $\hat{P}_N$, the empirical distribution of training data. If $\Omega$ is selected judiciously such that it is able to capture the test data distribution (under reasonable levels of perturbation), then the solution $\theta_\mathrm{DRO}$ is guaranteed to have robust out-of-sample performance. Meanwhile, we want to make $\Omega$ small enough to exclude irregular distributions that are not representative of the test data and incentivize over-conservative results. Note that DRO reduces to ERM when $\Omega$ shrinks to a singleton $\hat{P}_N$.

Previous works have considered constructing $\Omega$ based on moment conditions \cite{Delage,Goh}, as well as probability distance measures such as $f$-divergence \cite{Namkoong,Duchi} and Wasserstein distance \cite{Soroosh,Esfahani,Boston,Sinha}. Although many of these works demonstrate appealing theoretical guarantees and computational tractability, most of them do not admit a distributed implementation as explained previously. To this end, we resort to the Wasserstein DRO framework in \cite{Sinha}, which not only admits a reformulation that is solvable in the distributed setting, but also provides certified robustness under moderate levels of distributional shifts. See the details in the next subsection.

\subsection{Wasserstein DRO}
\label{sec2-2}
In Wasserstein DRO, the ambiguity set $\Omega$ is chosen as a Wasserstein ball $B_{\rho}(\hat{P}_N)=\{Q:W_c(Q,\hat{P}_N)\leq\rho\}$ with $\hat{P}_N$ at the center and $\rho$ being the radius, and $W_c(\cdot,\cdot)$ is the Wasserstein distance between two probability distributions with $c(\cdot,\cdot)$ being the transportation cost between two data points. Following a duality result, \cite{Sinha} proves the equality $\sup_{Q\in B_{\rho}(\hat{P}_N)}\mathbb{E}_{x\sim Q}f(\theta;x)=\inf_{\lambda\ge0}\{\lambda\rho+\mathbb{E}_{x\sim \hat{P}_N}\phi_\lambda(\theta;x)\}$, where $\phi_\lambda(\theta;x)=\sup_z\{f(\theta;z)-\lambda c(z,x)\}$ represents the robust surrogate of $f(\theta;x)$ and $\lambda$ is the dual variable (see Proposition 1 therein). Giving up the ambitious goal of solving the original problem exactly with a prespecified $\rho$, i.e.,
\begin{equation}
\min_\theta\sup_{Q:W_c(Q,\hat{P}_N)\leq\rho}\mathbb{E}_{x\sim Q}f(\theta;x)
\end{equation}
\cite{Sinha} instead seeks to solve an easier problem \begin{equation}
\min_\theta\mathbb{E}_{x\sim \hat{P}_N}\phi_\lambda(\theta;x)
\end{equation} with a fixed $\lambda\ge0$. \cite{Sinha} provides a certificate of robustness for any $\rho$ in (1) (which means any $\lambda$, by duality), justifying the switch from (1) to (2). In this way, the original infinite-dimensional optimization problem (1) is transformed into a tractable ERM-like problem (2). Moreover, since the objective function in (2) is simply the average of a cluster of empirical losses each defined by a single sample $x$ in the training set, (2) immediately admits a distributed implementation where each worker can calculate gradient-like updates based on its own local data. This attribute differentiates (2) from other centralized DRO reformulations and makes it suitable for our work.

Although \cite{Sinha} is a frequently cited paper in the field of machine learning, few works try to take advantage of the distributed nature of (2) and implement it in the distributed/federated learning setting. The only exceptions are \cite{DRFL} and \cite{DRFL2}, with the former proposing a framework named Distributionally Robust Federated Learning (DRFL) and the latter proposing a two-stage attack strategy to jeopardize the performance of DRFL. However, neither of these works establish certified robustness against both distributional shifts and byzantine attacks as does in this paper.

\section{Norm-based Screening (NBS)}
\label{sec3}

As a robust aggregation measure, the idea of NBS is fairly simple: leave out the vector inputs (i.e., local updates) with large norms and take the average of the remaining inputs as output. In this way, the influence of an erroneous/malicious input is properly bounded: it either is filtered out for having a large norm or can only finitely impact the output with a norm comparable to some benign inputs. We formally define NBS as a function “$\mathbf{Norm\_Screen}$”, as detailed in Algorithm 1.

\begin{algorithm}
	\caption{Norm-based Screening}\label{alg1}
	\renewcommand{\algorithmicrequire}{\textbf{Input:}}
	\renewcommand{\algorithmicensure}{\textbf{Output:}}
	\begin{algorithmic}[1]
		\REQUIRE
		$g_1,\ldots,g_m$ ($m$ vector inputs), screening percentage  $\beta$
		\ENSURE
		$G=\mathbf{Norm\_Screen}_\beta(g_1,\ldots,g_m)$
		\STATE  generate a new set of indices $(1),\ldots,(m)$, such that $\|g_{(1)}\|\le\cdots\le\|g_{(m)}\|$
		\STATE define an index set $\mathcal{U}=\{(1),\ldots,((1-\beta)m)\}$, which specifies the unscreened inputs
		\STATE calculate the output by averaging the unscreened inputs $G=\frac{1}{|\mathcal{U}|}\sum_{i\in\mathcal{U}}g_i$
	\end{algorithmic}
\end{algorithm}

Although NBS has previously been applied to screen byzantine-prone local gradients \cite{ITA,Ghosh,ISIT} and Newton updates \cite{newton}, we argue that NBS did not get its fair share of appreciation and publicity, partly because its robust property has not been formally stated and theorized. To fill this gap, we formulate an important property of NBS as explicated in Theorem 1, whose proof is deferred to Appendix \ref{pend_A}.
\begin{theorem}
	Suppose that a percentage of $\alpha\le\frac{1}{2}$ among $m$ inputs $g_1,\ldots,g_m$ are byzantine, whose indices compose a set $\mathcal{B}$ ($|\mathcal{B}|=\alpha m$), and the index set of honest inputs is denoted as $\mathcal{M}$ ($|\mathcal{M}|=(1-\alpha) m$). With $G=\mathbf{Norm\_Screen}_\beta(g_1,\ldots,g_m)$ and $\beta\ge\alpha$, the following inequality holds:
	\begin{equation}\label{theo_1}
	\|G-S\|\le\frac{2\alpha}{1-\beta}\|S\|+\max_{i\in\mathcal{M}}\|g_i-S\|
	\end{equation}
	where $S$ can be any vector with the same dimension as $G$.
\end{theorem}

Theorem 1 plays an essential role in the convergence analysis of our algorithm, as it properly upper-bounds the distance between the robustly aggregated gradient and the true global gradient without making any unjustified assumptions on the distribution of local gradients (see Lemma 2 in Section \ref{sec6-1}). In addition, as shown in Section \ref{sec6-3}, our intuition on the breakpoint of NBS is drawn from Lemma 2, which is credited to the explicit exposition of Theorem 1.

We note that a similar result to (3) has appeared in the existing work \cite{ITA} as an intermediate step in the derivation, though being mixed with other terms (see Section 9.1 therein). While giving \cite{ITA} its due credit, we argue that a more formal statement of this property is well-deserved.

\section{Problem Statement}
\label{sec4}
In this section, we formulate the problem of robust distributed learning under both distributional shifts and byzantine attacks.

\textbf{Basic setting.}
We consider a typical distributed learning scenario with one central server and $m$ parallel workers, among whom a total of $N$ data points $x_1,\ldots,x_N$ are allocated/collected for training. For simplicity and clear exposition, we assume an even data-split scenario where worker $i$ holds $n$ samples $x_{(i-1)n+1},\ldots,x_{(i-1)n+n}$ for $i=1,\ldots,m$ with $mn=N$. Note that uneven data-split cases can easily fit into our framework with minor adjustment.

\textbf{Learning goal.}
Let $f(\theta;x_j)$ be the loss function contingent upon model parameter $\theta$ and sample $x_j$. We aim for a model that has robust performance on the test data, which may exhibit some degree of distributional shifts from the training data.  According to the discussion in Section \ref{sec2-2}, such a model can be acquired by solving (2), whose solution enjoys theoretically-proven robustness against data perturbations. Specifically, we seek to minimize the objective $F(\theta)=\frac{1}{N}\sum_{j=1}^{N}\phi_\lambda(\theta;x_j)$ in which $\phi_\lambda(\theta;x_j)=\sup_z\{f(\theta;z)-\lambda c(z,x_j)\}$ is the robustified version of $f(\theta;x_j)$.

\textbf{Byzantine attack.}
We assume that a percentage $\alpha$ of local workers are byzantine and the remaining $1-\alpha$ are normal/honest. The sets of byzantine workers and honest workers are denoted as $\mathcal{B}$ and $\mathcal{M}$ respectively, with $|\mathcal{B}|=\alpha m$ and $|\mathcal{M}|=(1-\alpha) m$. During each training iteration, the server would ask all workers to conduct certain computational task based on their respective local data and to report the  result back to the server. While honest workers would follow the given instructions faithfully, byzantine workers need not to obey the protocol and can send arbitrary messages to the server. By convention, we assume that byzantine workers have complete knowledge of the system and learning algorithms, which allows them to generate the most damaging updates to attack the system.

\section{Proposed Algorithm}
\label{sec5}
On the macro level, our algorithm is based on distributed gradient descent combined with robust aggregation, through the following three key components.

\textbf{Gradient computation.}
We first consider a single unit of the objective function, i.e., $\phi_\lambda(\theta;x_j)$. To calculate its gradient on a fixed model $\theta_t$, \cite{Sinha} proposes to first find the maximizer, i.e., $z_j^*(\theta_t)=\arg\sup_z\{f(\theta_t;z)-\lambda c(z,x_j)\}$, and then take the gradient of $f(\theta;z_j^*(\theta_t))$ before replacing $\theta$ with $\theta_t$. The correctness of this approach is guaranteed by the following equation:
\begin{equation}
\begin{split}
\nabla_\theta\phi_\lambda(\theta;x_j)|_{\theta=\theta_t}&=\nabla_\theta\big[f(\theta;z_j^*(\theta_t))-\lambda c(z_j^*(\theta_t),x_j)\big]_{\theta=\theta_t}\\
&=\nabla_\theta f(\theta;z_j^*(\theta_t))|_{\theta=\theta_t}.
\end{split}
\end{equation}

To simplify notations, we denote $\nabla_\theta f(\theta;z)|_{\theta=\theta_t}$ as $\nabla_\theta f(\theta_t;z)$ where $z$, sometimes taking the form of $z(\theta_t)$, is always treated as a constant in the differentiation step.

\textbf{$\varepsilon$-approximation.} In most cases, the maximizer $z_j^*(\theta_t)$ does not have a closed-form solution, and thus can only be solved to a certain precision via iterative methods. Therefore, we only require workers to obtain an $\varepsilon$-optimal maximizer $z_j^\varepsilon(\theta_t)$, satisfying $\|z_j^\varepsilon(\theta_t)-z_j^*(\theta_t)\|\le\varepsilon$. This approximation offers a tradeoff between computational cost and model accuracy. In the next section, we will analyze both the effects of $\varepsilon$-approximation on model convergence and the cost of obtaining such an $\varepsilon$-optimal maximizer.

\textbf{Robust aggregation.} After obtaining the $\varepsilon$-optimally perturbed samples, each honest worker computes its (approximate) local gradient before sending it to the server, while byzantine workers would craft their own ill-intended gradients (denoted as $\star$). On the other end, the server robustly aggregates the received local gradients via NBS and uses the result to update the model. Here we assume that the proportion $\alpha$ of byzantine workers is known, and we always enforce that $\beta\ge\alpha$.

The detailed procedure of our algorithm is given in Algorithm 2.

\begin{algorithm}
	\caption{Distributional \& Byzantine Robust Distributed Gradient Descent}\label{alg2}
	\renewcommand{\algorithmicrequire}{\textbf{Input:}}
	\renewcommand{\algorithmicensure}{\textbf{Output:}}
	\begin{algorithmic}[1]
		\REQUIRE screening percentage $\beta$ ($\ge\alpha$), learning rate $\eta$, model initialization $\theta_0$, total iteration $T$
		\ENSURE completed model $\theta_T$
		\FOR{$t=0,1,\ldots,T-1$}
		\STATE \textbf{Server}: send $\theta_t$ to all workers
		\FOR{$i=1,2,\ldots,m$}
		\STATE \textbf{Worker $i$}: receive model $\theta_t$ from the server
		\STATE
		obtain $z_j^\varepsilon(\theta_t)$ for each local sample $x_j$ by solving $\sup_z\{f(\theta_t;z)-\lambda c(z,x_j)\}$ to $\varepsilon$-precision
		\STATE
		compute local gradient $g_i(\theta_t)=\left\{\begin{array}{cc}
		\frac{1}{n}\sum_{j=(i-1)n+1}^{(i-1)n+n}\nabla_\theta f(\theta_t;z_j^\varepsilon(\theta_t)) &\ i\in\mathcal{M}\\
		\star &i\in\mathcal{B}
		\end{array}
		\right.$
		\STATE
		send $g_i(\theta_t)$ to the server		
		\ENDFOR
		\STATE
		\textbf{Server}:
		collect $g_1(\theta_t),\ldots,g_m(\theta_t)$ from the workers
		\STATE
		compute the aggregated gradient $G(\theta_t)=\mathbf{Norm\_Screen}_\beta(g_1(\theta_t),\ldots,g_m(\theta_t))$
		\STATE
		update model $\theta_{t+1}=\theta_t-\eta\cdot G(\theta_t)$
		\ENDFOR
	\end{algorithmic}
\end{algorithm}

\section{Convergence Analysis}

\subsection{Preliminaries}\label{sec6-1}
To delineate the convergence behavior of the proposed algorithm, we adopt some widely used assumptions as below. Assumptions 1-3 concern the distributional shifts as in \cite{Sinha}, which hold for tractable scenarios. Assumption 4 upper-bounds the distance between the average gradient and the gradient w.r.t. a single sample, which is characteristic of gradient averaging methods and clearly holds.

\textit{Assumption 1:} The loss function $f(\theta;z)$ satisfies the Lipschitzian smoothness conditions
\begin{equation*}
\begin{split}
\|\nabla_\theta f(\theta_1;z)-\nabla_\theta f(\theta_2;z)\|&\le L_{\theta\theta}\|\theta_1-\theta_2\|,\\
\|\nabla_\theta f(\theta;z_1)-\nabla_\theta f(\theta;z_2)\|&\le L_{\theta z}\|z_1-z_2\|,\\
\|\nabla_z f(\theta_1;z)-\nabla_z f(\theta_2;z)\|&\le L_{z\theta}\|\theta_1-\theta_2\|,\\
\|\nabla_z f(\theta;z_1)-\nabla_z f(\theta;z_2)\|&\le L_{zz}\|z_1-z_2\|.
\end{split}
\end{equation*}

\textit{Assumption 2:} The function $c(z,x)$ defined in the Wasserstein metric is $L_c$-smooth and 1-strongly convex w.r.t. $z$.

\textit{Assumption 3:} The dual variable $\lambda$ satisfies $\lambda>L_{zz}$ where $L_{zz}$ is defined in Assumption 1.

\textit{Assumption 4:} For any specific $\theta_t$, it holds that
\begin{equation}\label{main_1}
\max_{1\le k\le N}\left\|\nabla_\theta f(\theta_t;z_k^*(\theta_t))-\frac{1}{N}\sum_{j=1}^{N}\nabla_\theta f(\theta_t;z_j^*(\theta_t))\right\|\le\sigma.
\end{equation}

Based on the above assumptions, we formulate two lemmas that will serve as core building blocks of the ensuing theorems on convergence. We should note that Lemma 1 is a direct result of \cite{Sinha} (see Lemma 1 therein). For completeness, we summarize the proof of Lemma 1 in Appendix \ref{pend_B}, matching the notations of this paper.

\begin{lemma}
	Under Assumptions 1-3, the objective function $F(\theta)=\frac{1}{N}\sum_{j=1}^{N}\phi_\lambda(\theta;x_j)$ is $L_F$-smooth with $L_F=L_{\theta\theta}+\frac{L_{\theta z}L_{z\theta}}{\lambda-L_{zz}}$.
\end{lemma}

Lemma 1 specifies the smoothness level of the objective function, thus allowing standard gradient descent to make steady progress with a proper step size, such as $\frac{1}{L_F}$. However, in our problem, the error-free gradient is unattainable due to the byzantine nodes. To this end, we propose Lemma 2 that quantifies the deviation of our implemented gradient from the true global gradient $\nabla F(\theta_t)$. The proof of Lemma 2 is deferred to Appendix \ref{pend_C}.

\begin{lemma}
	Under Assumptions 1-4, for any specific $\theta_t$, it holds that
	\begin{equation}\label{lem_2}
	\|G(\theta_t)-\nabla F(\theta_t)\|\le\frac{2\alpha}{1-\beta}\|\nabla F(\theta_t)\|+(L_{\theta z}\varepsilon+\sigma)
	\end{equation}
	where $G(\theta_t)$ is the aggregated gradient in Algorithm 2 (line 10).
\end{lemma}

\subsection{Main Theorems}
\label{sec6-2}
\subsubsection{Nonconvex Losses}
We first consider the most general case of the loss function $f(\theta;z)$ being nonconvex in $\theta$, such as in neural network training. For this case, we derive Theorem 2 that guarantees convergence of our algorithm to a stationary point of the objective function. The proof of Theorem 2 is deferred to Appendix \ref{pend_D}.

\begin{theorem}
	Suppose that Assumptions 1-4 hold and $\alpha<\frac{1}{3}$. Taking $\eta=\frac{1}{L_F}$, Algorithm 2 satisfies
	\begin{equation}\label{theo_2}
	\begin{split}
	\frac{1}{T}\sum_{t=0}^{T-1}\|\nabla F(\theta_t)\|^2\le&\frac{2L_F}{\big(1-(1+r)C_\alpha^2\big)T}\big[F(\theta_0)-F(\theta^*)\big]\\
	&+\frac{(1+1/r)(L_{\theta z}\varepsilon+\sigma)^2}{1-(1+r)C_\alpha^2}
	\end{split}	
	\end{equation}
	where $\theta^*$ is the global minimizer of $F(\theta)$, $C_\alpha=\frac{2\alpha}{1-\beta}$, and $r$ should satisfy $0<r<\big(\frac{1-\beta}{2\alpha}\big)^2-1$.
\end{theorem}

\subsubsection{Convex Losses}
Now we consider the case where the loss function is convex as in Assumption 5. Additionally, we make Assumption 6 suggesting that all the intermediate iterations would not be infinitely worse than the initialization $\theta_0$. We propose Theorem 3 that grants our algorithm convergence guarantee in the convex regime. The proof of Theorem 3 is deferred to Appendix \ref{pend_E}.

\textit{Assumption 5:} The loss function $f(\theta;z)$ is convex w.r.t. $\theta$.

\textit{Assumption 6:}  There exists a fixed $k$ such that $\|\theta_t-\theta^*\|\le k\|\theta_0-\theta^*\|$ holds for $t=0,1,\ldots,T-1$.

\begin{theorem}
	Suppose that Assumptions 1-6 hold and $\alpha<\frac{1}{3}$. Taking $\eta=\frac{1}{L_F}$, Algorithm 2 satisfies
	\begin{equation}\label{theo_3}
	\begin{split}
	&F(\theta_T)-F(\theta^*)\le\max\left\{
	\frac{4L_FD^2}{\big(1-(1+r)C_\alpha^2\big)T},\right.\\
	&\left.\sqrt{\frac{2(1+1/r)}{1-(1+r)C_\alpha^2}}D(L_{\theta z}\varepsilon+\sigma)+\frac{(1+1/r)(L_{\theta z}\varepsilon+\sigma)^2}{2L_F}\right\}
	\end{split}	
	\end{equation}
	where $D=k\|\theta_0-\theta^*\|$, $C_\alpha$, $r$ are the same as in Theorem 2.
\end{theorem}

\subsubsection{Strongly Convex Losses}
Finally, we assume strong convexity on the objective function as in Assumption 7, in which case we propose Theorem 4 that guarantees convergence of our algorithm to the optimal model $\theta^*$. The proof of Theorem 4 is deferred to Appendix \ref{pend_F}.

\textit{Assumption 7:} The objective function $F(\theta)$ is $\lambda_F$-strongly convex.

\begin{theorem}
	Suppose that Assumptions 1-4 and 7 hold, and $\alpha<\frac{1}{1+2L_F/\lambda_F}<\frac{1}{3}$. Taking $\eta=\frac{2}{L_F+\lambda_F}$, Algorithm 2 satisfies (with $C_\alpha=\frac{2\alpha}{1-\beta}$)
	\begin{equation}\label{theo_4}
	\begin{split}
	\|\theta_T-\theta^*\|\le&\left(\frac{2L_FC_\alpha+L_F-\lambda_F}{L_F+\lambda_F}\right)^T\|\theta_0-\theta^*\|\\
	&+\frac{L_{\theta z}\varepsilon+\sigma}{\lambda_F-L_FC_\alpha}.
	\end{split}	
	\end{equation}
\end{theorem}

\noindent
\textbf{Observations.}
According to Theorems 2-4, Algorithm 2 is able to achieve some sense of convergence under all three cases. Meanwhile, we can clearly identify the effects of byzantine percentage $\alpha$ and suboptimality level $\varepsilon$ on convergence: a larger $\alpha$ (entailing larger $C_\alpha$) not only decreases convergence speed, but also increases convergence error, whereas $\varepsilon$ only affects the convergence error and has no impact on the convergence rate.

\subsection{The Breakpoint of NBS}\label{sec6-3}
We define the breakpoint of a certain algorithm as the minimum byzantine percentage at which that algorithm cannot converge. According to Theorems 2-4, the breakpoint of our algorithm is $\frac{1}{3}$. (Although Theorem 4 requires that $\alpha<\frac{1}{1+2L_F/\lambda_F}$, it can still converge as in (\ref{theo_2}) under $\alpha<\frac{1}{3}$ by taking $\eta=\frac{1}{L_F}$.) In fact, based on the derivations in Appendices \ref{pend_D} and \ref{pend_F}, we assert that for any algorithm that incorporates NBS to converge, it always should hold that $\alpha<\frac{1}{3}$. This insight can be drawn from Lemma 2, where the distance between $G(\theta_t)$ and $\nabla F(\theta_t)$ is upper-bounded by two terms. In the convergence proofs, we found that the coefficient of the first term must be less than $1$, i.e., $\frac{2\alpha}{1-\beta}<1$, which, combined with $\beta\ge\alpha$, imposes that $\alpha<\frac{1}{3}$.

We also notice that previous works implementing NBS for byzantine robustness uniformly claimed that the breakpoint of their algorithms is $\frac{1}{2}$ \cite{ITA,Ghosh,ISIT,newton}. The fallacy of this claim can be illustrated by a simple counterexample: suppose that there are $4$ byzantine updates $g_1,g_2,g_3,g_4$ and $6$ honest updates $g_5,g_6,\ldots,g_{10}$ (in norm-descending order). If the byzantine updates are crafted such that $g_1=g_2=g_3=g_4=-g_9$, then the NBS output with $\beta=\alpha=0.4$ is totally dominated by byzantine updates as $G=\frac{g_1+g_2+g_3+g_4+g_9+g_{10}}{6}$. If this happens for every iteration, then the algorithm surely would not converge to the correct solution.

To the best of our knowledge, our claim that the breakpoint of NBS is $\frac{1}{3}$ is new in the literature.

\subsection{Discussion}
\label{sec6-4}
\textbf{The cost of computing $\varepsilon$-optimal maximizer.}
To obtain each perturbed sample $z_j^\varepsilon(\theta_t)$, we need to maximize $g(z)=f(\theta_t,z)-\lambda c(z,x_j)$ (at fixed $\theta_t$ and $x_j$) to $\varepsilon$-optimality. It turns out that $g(z)$ is both smooth and strongly concave, which, according to optimization theory, suggests that maximizing $g(z)$ enjoys linear convergence rate using gradient method. Specifically, according to Assumption 1, $-L_{zz}\cdot\mathbf{I}\preceq\nabla_z^2 f(\theta;z)\preceq L_{zz}\cdot\mathbf{I}$; according to Assumption 2, $1\cdot\mathbf{I}\preceq\nabla_z^2 c(z,x)\preceq L_c\cdot\mathbf{I}$; therefore, $-(\lambda L_c+L_{zz})\cdot\mathbf{I}\preceq \nabla_z^2 g(z)\preceq -(\lambda-L_{zz})\cdot\mathbf{I}$, which means that $g(z)$ is $L_g$-smooth and $\lambda_g$-strongly concave with $L_g=\lambda L_c+L_{zz}$ and $\lambda_g=\lambda-L_{zz}$. According to convergence analysis similar to that of Appendix \ref{pend_F}, by iterating $z_{t+1}=z_t+\eta_z\nabla g(z_t)$ with $\eta_z=\frac{2}{L_g+\lambda_g}=\frac{2}{\lambda L_c+\lambda}$, we have $\|z_T-z^*\|\le p^T\|z_0-z^*\|$ with the exact maximizer $z^*$ and the convergence factor $p=\frac{L_g-\lambda_g}{L_g+\lambda_g}=\frac{2L_{zz}+\lambda L_c-\lambda}{\lambda L_c+\lambda}$. As a result, to obtain an $\varepsilon$-optimal solution $z^\varepsilon$ satisfying $\|z^\varepsilon-z^*\|\le\varepsilon$ requires that $T_z\ge\frac{\ln(D_z/\varepsilon)}{\ln(1/p)}$ where $T_z$ is the number of gradient ascent iterations and $D_z=\|z_0-z^*\|$. Here we show that $\varepsilon$ is easily adjustable by tuning $T_z$, and a smaller $\varepsilon$ only entails a moderate increase in $T_z$.

\textbf{The strategy of adjusting $\varepsilon$.}
According to our analysis, small $\varepsilon$ corresponds to low model error. On the other hand, enforcing small $\varepsilon$ puts relatively heavy computational workload on local workers. To achieve a good tradeoff between computational cost and model accuracy, we recommend a two-stage strategy where a big $\varepsilon$ is adopted in the beginning stage of training and a small $\varepsilon$ is enforced in the ending stage. This is because in the beginning stage the byzantine gradients would contaminate the aggregated gradient to a large degree, and there is no need for honest workers to calculate their perturbed data/local gradients with very high precision. To make it clearer, we refer to (\ref{lem_2}) where $\frac{2\alpha}{1-\beta}\|\nabla F(\theta_t)\|$ is the dominant term at the beginning and therefore a relatively big $\varepsilon$ would have little impact on the converging process. In the ending stage where $\nabla F(\theta_t)$ approaches zero, $L_{\theta z}\varepsilon+\sigma$ becomes the dominant term, and we switch into a small $\varepsilon$ regime to achieve high model accuracy, at the cost of concentrated computation in the end.

\textbf{The effects of non-iid data.}
Recall that in Section \ref{sec4} where the targeted problem is formulated, we did not assume that the data are iid across all workers, suggesting that our convergence results should hold in non-iid cases as well. This is due to the analytical approach we take in the proof of Lemma 2, where we bound the maximal distance between honest local gradients and the targeted gradient as $\max_{i\in\mathcal{M}}\|g_i(\theta_t)-\nabla F(\theta_t)\|\le L_{\theta z}\varepsilon+\sigma$, regardless of local data distribution. In practice, this distance should increase if the distribution of local data goes from iid to highly pathological/non-iid. However, for the convenience of analysis, this subtle difference is erased through the adoption of a universal upper-bound $L_{\theta z}\varepsilon+\sigma$. There are two major takeaways from this observation. On one hand, one should be aware that our theoretical results may not reflect the empirical effects of non-iid data on the convergence, since our emphasis is on the effects of $\alpha$ and $\varepsilon$. On the other hand, this analytical approach we adopted, i.e., making assumptions in the spirit of Assumption 4 and imposing a universal upper-bound to eliminate local updates' differences, might serve as a pathway for future works to bypass the non-iid issue theoretically in the convergence analysis.

\section{Simulations}

\subsection{Setup}

\textbf{Learning model and dataset.} In this section, we empirically evaluate the effectiveness of our algorithm
for a classification task using logistic regression model on the Spambase dataset \cite{spambase}. Of the total $4601$ email messages, we use $\frac{2}{3}$ for training and $\frac{1}{3}$ for testing, with the same ratio of spams to non-spams in two sets. The training data are evenly split across $m=20$ workers. Upon completion of training, we evaluate the misclassification rate on the test set as the performance metric.

\textbf{Shift model.}
To simulate distributional shifts, we put a certain perturbation on test data with a controlled budget $q$ under both $L_1$ and $L_2$ norm. Since in supervised learning it is a common practice to only perturb the feature vector $x$, not the label $y$, we perturb each test data $(x,y)$ into $(z,y)$ satisfying $\|z-x\|_p\le q$ ($p=1,2$) and the $z$'s are chosen to maximally increase the cross-entropy loss function on test data.

\textbf{Byzantine model.}
We experiment with two types of byzantine attacks. In the first case, the byzantine gradient is $-10g$ where $g$ is the true global gradient. Such attack is totally destructive to simple averaging but can be easily filtered out by most robust aggregation measures due to its conspicuously large norm. In the second case, the byzantine gradient is $\big(0.8\|g\|\big)\cdot h$, where $h$ is a random Gaussian vector scaled to unit-norm, i.e., $\|h\|=1$. This kind of attack is not as lethal as the former to simple averaging, but can bypass and thus exhibit full impact on NBS because its norm is not among the largest of local gradients, but still significant enough to impact the aggregated result. We refer to these two types of attacks as aggressive attack and intelligent attack respectively.

\textbf{Data perturbation in training.} Our algorithm requires honest workers to obtain each perturbed sample by approximately solving $\sup_z\{f(\theta;z)-\lambda c(z,x)\}$, in which we set $c(z,x)=\frac{1}{2}\|z-x\|^2$ in order to satisfy Assumption 2 (as in \cite{Sinha}). In accordance with the adversarial perturbation on test data, we only perturb the feature vector $x$ into $z$ without changing the label $y$. For logistic regression, the objective is $g(z)=-y\ln a-(1-y)\ln(1-a)-\frac{\lambda}{2}\|z-x\|^2$ with $a=1/(1+e^{-\theta^\mathrm{T}z})$, which has no closed-form solution. Therefore, we calculate the approximate maximizer via gradient ascent $z_{t+1}=z_{t}+\eta_z\nabla g(z_t)$ using $T_z$ iterations initialized at $z_0=x$. Throughout the experiments, we set $\lambda=3$, $\eta_z=0.05$, and $T_z=10$ (if not otherwise specified).

\subsection{Evaluation}
First, we compare the performances of four algorithms: \textbf{Algorithm 2} with $\beta m=3$ screened gradients, \textbf{DRO} (replacing NBS with averaging in Algorithm 2), \textbf{NBS} (skipping data perturbation in Algorithm 2), and \textbf{ERM} (standard distributed gradient descent). Since Algorithm 2 is the only algorithm that is designed to solve our proposed problem at this point, we use the latter three algorithms as benchmarks. These four algorithms are tested in five environments listed below (E0--E4), with step size $\eta=1$, iterations $T=300$, and shared random initialization. We summarize the learning results (misclassification rate) in Table I.
\begin{itemize}
	\item[E0:] no byzantine attack ($\alpha m=0$) \& no data shift ($q=0$)
	\item[E1:] aggressive attack with $\alpha m=3$ \& $L1$ shift with $q=0.3$
	\item[E2:] aggressive attack with $\alpha m=3$ \& $L2$ shift with $q=0.3$
	\item[E3:] intelligent attack with $\alpha m=3$ \& $L1$ shift with $q=0.3$
	\item[E4:] intelligent attack with $\alpha m=3$ \& $L2$ shift with $q=0.3$	
\end{itemize}

\begin{table}[h]
	\label{tab1}
	\caption{Comparison of 4 Algorithms.}
	\renewcommand\arraystretch{1.5}
	\centering
	\setlength{\tabcolsep}{2.5mm}{
	\begin{tabular}{|c|c|c|c|c|}
		\hline
		& ERM    & NBS    & DRO    & Alg.   2 \\ \hline
		Environment   0 & 0.0959 & 0.1076 & 0.0926 & 0.1037        \\ \hline
		Environment   1 & 0.4658 & 0.1931 & 0.4697 & 0.1350        \\ \hline
		Environment   2 & 0.4866 & 0.3170 & 0.4912 & 0.2322        \\ \hline
		Environment   3 & 0.1931 & 0.2107 & 0.1357 & 0.2048        \\ \hline
		Environment   4 & 0.3138 & 0.3164 & 0.2290 & 0.2779        \\ \hline
	\end{tabular}}
\end{table}
From Table I, we observe that aggressive attack favors screening algorithms Algorithm 2 and NBS, and disfavors non-screening algorithms DRO and ERM, while intelligent attack is the opposite. One the other hand, DRO and ERM totally fail under aggressive attack, but Algorithm 2 and NBS achieve degraded but still acceptable performance under intelligent attack, which makes Algorithm 2 and NBS safer choices under byzantine attacks in general. Since Algorithm 2 is also built with distributional robustness, we can see that it outperforms NBS in all tested scenarios.

Next, we focus on Algorithm 2 and test its performance under different levels of shifts or byzantine attacks. In Figure 1, we plot the performance curves of Algorithm 2 and NBS under aggressive attack with $\alpha m=3$ and $L1$ shift with varying $q$ (we exclude DRO and ERM because they cannot endure aggressive attack as shown in Table I). From Figure 1, we can see that  the misclassification rate of Algorithm 2 increases much slowlier than that of NBS as $q$ increases, exhibiting our algorithm's effectiveness against varying level of distributional shifts. In Figure 2, we plot the performance curves of Algorithm 2 and DRO under $L1$ shift with $q=0.3$ and varying number of byzantine nodes $\alpha m$. We experiment with both aggressive attack and intelligence attack and choose the worse result for each algorithm (since it is not fair to test only one type of attack). From Figure 2, we can see that Algorithm 2 offers good protection against byzantine attacks when $\alpha m\le\beta m$, but would collapse if $\alpha>\beta$.

\begin{figure}[t]
	\centering
	\includegraphics[width=3.0in]{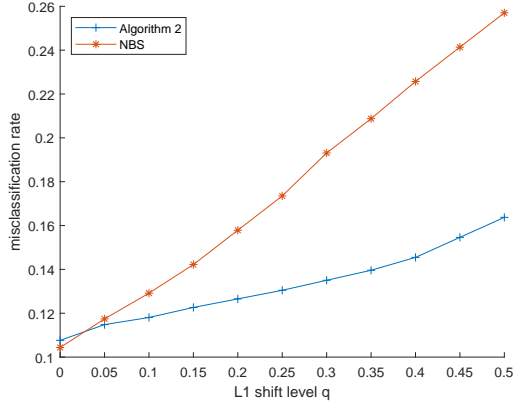}
	\caption{Performance curves under different level of shifts.}
\end{figure}
\begin{figure}[t]
	\centering
	\includegraphics[width=3.0in]{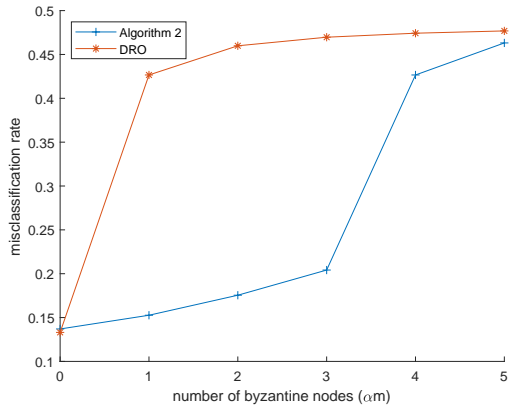}
	\caption{Performance curves under different number of byzantine nodes.}
\end{figure}

Finally, we explore the influence of parameter selection on Algorithm 2 under fixed attacks and shifts (we choose Environment 1). In Figure 3, we test different values of $\lambda$, which is the fixed dual variable that has to be selected empirically according to \cite{Sinha}. Here we set $T_z=100$ to make the curve smooth. From Figure 3, we can see that the performance of Algorithm 2 is quite stable with different values of $\lambda$ (as long as $\lambda$ is not overly small). In Figure 4, we test different numbers of $T_z$. From Figure 4, we can see that to achieve good performance does not require too many iterations ($T_z=10$ suffices in this case). This result corroborates the first remark in Section \ref{sec6-4} that computing perturbed samples with high precision only requires a moderate increase in the computational cost. Figure 3 and Figure 4 suggest that our algorithm's effectiveness is not sensitive to the selection of hyper-parameters, which is a desirable attribute in practice.

\begin{figure}[t]
	\centering
	\includegraphics[width=3.0in]{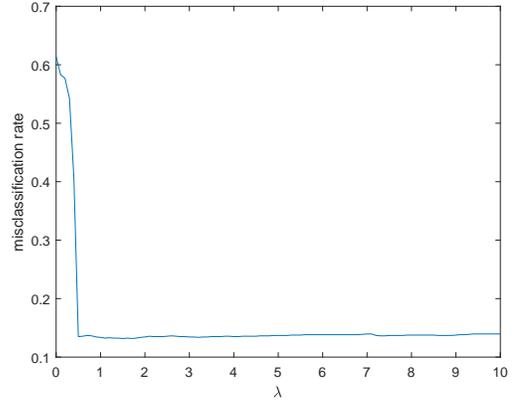}
	\caption{The performance of Algorithm 2 with different $\lambda$.}
\end{figure}
\begin{figure}[t]
	\centering
	\includegraphics[width=3.0in]{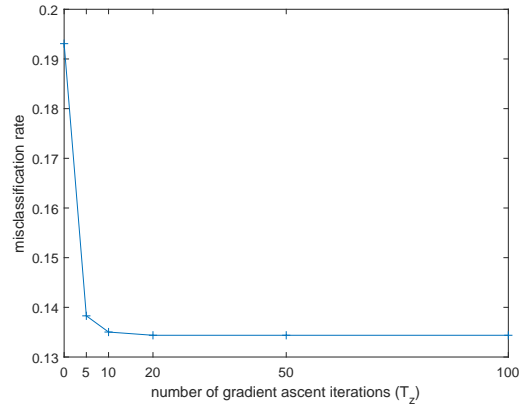}
	\caption{The performance of Algorithm 2 with different $T_z$.}
\end{figure}

\section{Conclusion}
In this paper, we address the uncharted problem of robust distributed learning at the presence of both distributional shifts and byzantine attacks. We propose a new algorithm that incorporates effective robust features to defend against both safety threats. The convergence of the proposed algorithm is theoretically guaranteed for different types of learning models. We also empirically demonstrate that our algorithm enjoys satisfying performance, matching the theoretical results.

\appendices
\setlength\parindent{0pt}

	\section{Proof of Theorem 1}
	\label{pend_A}
	For $G=\frac{1}{|\mathcal{U}|}\sum_{i\in\mathcal{U}}g_i$ with $\mathcal{U}=\{(1),\ldots,((1-\beta)m)\}$ and any specific vector $S$, we have
	\begin{equation*}
	\begin{split}
	\|G-S\|=&\left\|\frac{1}{|\mathcal{U}|}\sum_{i\in\mathcal{U}}g_i-S\right\|\\
	=&\left\|\frac{1}{|\mathcal{U}|}\sum_{i\in\mathcal{U}}(g_i-S)\right\|\\
	=&\frac{1}{|\mathcal{U}|}\left\|\sum_{i\in\mathcal{U}\cap\mathcal{M}}(g_i-S)+\sum_{i\in\mathcal{U}\cap\mathcal{B}}(g_i-S)\right\|\\
	\le&\frac{1}{|\mathcal{U}|}\left(\sum_{i\in\mathcal{U}\cap\mathcal{M}}\|g_i-S\|+\sum_{i\in\mathcal{U}\cap\mathcal{B}}\|g_i-S\|\right).
	\end{split}
	\end{equation*}
	For $i\in\mathcal{U}\cap\mathcal{M}$,  $\|g_i-S\|\le\Delta$ (we define $\Delta=\max_{i\in\mathcal{M}}\|g_i-S\|$).	
	For $i\in\mathcal{U}\cap\mathcal{B}$, we bound $\|g_i-S\|$ as
	\begin{equation*}
	\begin{split}
	\|g_i-S\|\le&\|g_i\|+\|S\|\\
	\le&\|g_{((1-\beta)m)}\|+\|S\|\\
	\le&\|g_{((1-\alpha)m)}\|+\|S\|\\
	\le&\max_{i\in\mathcal{M}}\|g_i\|+\|S\|\\
	=&\max_{i\in\mathcal{M}}\|g_i-S+S\|+\|S\|\\
	\le&\max_{i\in\mathcal{M}}\|g_i-S\|+2\|S\|\\
	=&\Delta+2\|S\|.
	\end{split}
	\end{equation*}
	Combining the above results, we have
	\begin{equation*}\label{A_1}
	\begin{split}
	\|G-S\|\le&\frac{1}{|\mathcal{U}|}\Big(|\mathcal{U}\cap\mathcal{M}|\cdot\Delta+|\mathcal{U}\cap\mathcal{B}|\cdot(\Delta+2\|S\|)\Big)\\
	=&\frac{1}{|\mathcal{U}|}\Big(|\mathcal{U}|\cdot\Delta+2|\mathcal{U}\cap\mathcal{B}|\cdot\|S\|\Big)\\
	=&\Delta+\frac{2|\mathcal{U}\cap\mathcal{B}|}{|\mathcal{U}|}\|S\|\\
	\le&\Delta+\frac{2\alpha}{1-\beta}\|S\|
	\end{split}
	\end{equation*}
	which is exactly the conclusion in Theorem 1. Note that the last inequality only holds on condition that $|\mathcal{B}|\le|\mathcal{U}|$, i.e., $\alpha\le1-\beta$, which, combined with $\beta\ge\alpha$, suggests that $\alpha\le\frac{1}{2}$.
	
	\textbf{Comment.} In the above analysis, we upper-bound the term $\|g_i-S\|$ by $\Delta$ for honest inputs, and by $\Delta+2\|S\|$ for byzantine inputs. The first bound clearly holds on account of the definition of $\Delta$. We now use a toy example to verify the tightness of the second bound. Suppose that $S=5$, $g_1=4$, $g_2=6$ and $g_3=-5.9$ with $\mathcal{M}=\{1,2\}$ and $\mathcal{B}=\{3\}$. For $i\in\mathcal{M}$, $\|g_i-S\|\le\Delta=1$; for $i\in\mathcal{B}$ ($i=3$), $\|g_i-S\|=10.9$, which is narrowly below $\Delta+2\|S\|=11$. In this example, the byzantine input $g_3$ is carefully chosen such that it is able to bypass the norm-based screening with $\beta=\frac{1}{3}$, and meanwhile exerts maximum adversarial perturbation on the outcome. This shows that our results accurately reflect the potential damage that byzantine workers can cause.
	
	\section{Proof of Lemma 1}
	\label{pend_B}
	Define $g(\theta;z)=f(\theta;z)-\lambda c(z,x)$ (we fix $\lambda$ and $x$, and view them as constants). Since $f(\theta;z)$ is $L_{zz}$-smooth w.r.t. $z$ (Assumption 1) and $c(z,x)$ is 1-strongly convex (Assumption 2), we have$$
	\nabla^2_z g(\theta;z)=\nabla^2_z f(\theta;z)-\lambda\cdot\nabla^2_z c(z,x)\preceq-(\lambda-L_{zz})\cdot\mathbf{I}$$
	which shows that $g(\theta;z)$ is $(\lambda-L_{zz})$-strongly concave w.r.t. $z$.
	
	For any $\theta_1$ and $\theta_2$, define $z_1^*=\arg\sup_z g(\theta_1;z)$ and $z_2^*=\arg\sup_z g(\theta_2;z)$. Apparently we have $\nabla_z g(\theta_1;z_1^*)=\nabla_z g(\theta_2;z_2^*)=0$. According to the strong concavity of $g(\theta;z)$, we can obtain the following two inequalities:
	\begin{equation}\label{B_1}
	\begin{split}
	g(\theta_1;z_1^*)\le g(\theta_1;z_2^*)&+\langle\nabla_z g(\theta_1;z_2^*),z_1^*-z_2^*\rangle\\
	&-\frac{\lambda-L_{zz}}{2}\|z_1^*-z_2^*\|^2,
	\end{split}	
	\end{equation}
	\begin{equation}\label{B_2}
	\begin{split}
	g(\theta_1;z_2^*)\le g(\theta_1;z_1^*)&+\langle\nabla_z g(\theta_1;z_1^*),z_2^*-z_1^*\rangle\\
	&-\frac{\lambda-L_{zz}}{2}\|z_2^*-z_1^*\|^2.
	\end{split}	
	\end{equation}
	Adding (\ref*{B_1}) and (\ref*{B_2}) together, we have
	\begin{equation*}
	\begin{split}
	&(\lambda-L_{zz})\|z_1^*-z_2^*\|^2\\
	&\le\langle\nabla_z g(\theta_1;z_2^*),z_1^*-z_2^*\rangle\\
	&=\langle\nabla_z g(\theta_1;z_2^*)-\nabla_z g(\theta_2;z_2^*),z_1^*-z_2^*\rangle\\
	&=\langle\nabla_z f(\theta_1;z_2^*)-\nabla_z f(\theta_2;z_2^*),z_1^*-z_2^*\rangle\\
	&\le\|\nabla_z f(\theta_1;z_2^*)-\nabla_z f(\theta_2;z_2^*)\|\cdot\|z_1^*-z_2^*\|\\
	&\le L_{z\theta}\|\theta_1-\theta_2\|\cdot\|z_1^*-z_2^*\|
	\end{split}
	\end{equation*}
	which leads to
	\begin{equation*}\label{B_3}
	\|z_1^*-z_2^*\|\le\frac{L_{z\theta}}{\lambda-L_{zz}}\|\theta_1-\theta_2\|
	\end{equation*}
	
	Recall that $\nabla_\theta \phi_\lambda(\theta;x)=\nabla_\theta f(\theta;z^*(\theta))$ (see (4)), we have
	\begin{equation*}\label{B_4}
	\begin{split}
	&\|\nabla_\theta \phi_\lambda(\theta_1;x)-\nabla_\theta \phi_\lambda(\theta_2;x)\|\\
	=&\|\nabla_\theta f(\theta_1;z_1^*)-\nabla_\theta f(\theta_2;z_2^*)\|\\
	=&\|\nabla_\theta f(\theta_1;z_1^*)-\nabla_\theta f(\theta_1;z_2^*)+\nabla_\theta f(\theta_1;z_2^*)-\nabla_\theta f(\theta_2;z_2^*)\|\\
	\le&\|\nabla_\theta f(\theta_1;z_1^*)-\nabla_\theta f(\theta_1;z_2^*)\|+\|\nabla_\theta f(\theta_1;z_2^*)-\nabla_\theta f(\theta_2;z_2^*)\|\\
	\le& L_{\theta z}\|z_1^*-z_2^*\|+L_{\theta\theta}\|\theta_1-\theta_2\|\\
	\le&\left(L_{\theta\theta}+\frac{L_{\theta z}L_{z\theta}}{\lambda-L_{zz}}\right)\|\theta_1-\theta_2\|.
	\end{split}
	\end{equation*}
	According to the definition of smoothness, $\phi_\lambda(\theta;x)$ is $L_F$-smooth w.r.t. $\theta$ with $L_F=L_{\theta\theta}+\frac{L_{\theta z}L_{z\theta}}{\lambda-L_{zz}}$. As a result, $F(\theta)=\frac{1}{N}\sum_{j=1}^{N}\phi_\lambda(\theta;x_j)$ is also $L_F$-smooth.
	
	\section{Proof of Lemma 2}
	\label{pend_C}
	According to Theorem 1, we have the following result by setting $S=\nabla F(\theta_t)$
	\begin{equation}
	\label{C_1}
	\|G(\theta_t)-\nabla F(\theta_t)\|\le\frac{2\alpha}{1-\beta}\|\nabla F(\theta_t)\|+\max_{i\in\mathcal{M}}\|g_i(\theta_t)-\nabla F(\theta_t)\|.
	\end{equation}
	According to Algorithm 2, for $i\in\mathcal{M}$, $g_i(\theta_t)=\frac{1}{n}\sum_{j=(i-1)n+1}^{(i-1)n+n}\nabla_\theta f(\theta_t;z_j^\varepsilon(\theta_t))$. Defining an auxiliary term $g^*_i(\theta_t)=\frac{1}{n}\sum_{j=(i-1)n+1}^{(i-1)n+n}\nabla_\theta f(\theta_t;z_j^*(\theta_t))$ (where $z_j^*(\theta_t)$ is the exact maximizer), we can bound the distance between $g_i(\theta_t)$ and $g^*_i(\theta_t)$ for $\forall i\in\mathcal{M}$ as
	\begin{equation}\label{C_3}
	\begin{split}
	\|&g_i(\theta_t)-g^*_i(\theta_t)\|\\
	\le&\max_{1\le j\le N}\|\nabla_\theta f(\theta_t;z_j^\varepsilon(\theta_t))-\nabla_\theta f(\theta_t;z_j^*(\theta_t))\|\\
	\le&L_{\theta z}\max_{1\le j\le N}\|z_j^\varepsilon(\theta_t)-z_j^*(\theta_t)\|\\
	\le&L_{\theta z}\varepsilon
	\end{split}
	\end{equation}
	in which the second inequality follows from Assumption 1 and the third inequality follows from the definition of $z_j^\varepsilon(\theta_t)$.
	
	Next, we have
	\begin{equation}
	\label{C_2}
	\begin{split}
	&\max_{i\in\mathcal{M}}\|g_i(\theta_t)-\nabla F(\theta_t)\|\\
	&=\max_{i\in\mathcal{M}}\|g_i(\theta_t)-g_i^*(\theta_t)+g_i^*(\theta_t)-\nabla F(\theta_t)\|\\
	&\le\max_{i\in\mathcal{M}}\big(\|g_i(\theta_t)-g_i^*(\theta_t)\|+\|g_i^*(\theta_t)-\nabla F(\theta_t)\|\big)\\
	&\le L_{\theta z}\varepsilon+\max_{i\in\mathcal{M}}\|g_i^*(\theta_t)-\nabla F(\theta_t)\|\\
	&\le L_{\theta z}\varepsilon\\
	&\quad+\max_{1\le k\le N}\left\|\nabla_\theta f(\theta_t;z_k^*(\theta_t))-\frac{1}{N}\sum_{j=1}^{N}\nabla_\theta f(\theta_t;z_j^*(\theta_t))\right\|\\
	&\le L_{\theta z}\varepsilon+\sigma
	\end{split}
	\end{equation}
	where the second inequality follows from (\ref*{C_3}) and the last inequality follows from Assumption 4.
	
	Finally, combining (\ref*{C_1}) and (\ref*{C_2}) leads to the conclusion in Lemma 2.
	
	\section{Proof of Theorem 2}
	\label{pend_D}
	According to Lemma 1, $F(\theta)$ is $L_F$-smooth. According to the property of smoothness, we have
	\begin{equation}
	\label{D_1}
	\begin{split}
	&F(\theta_{t+1})\le F(\theta_t)+\langle\nabla F(\theta_t),\theta_{t+1}-\theta_t\rangle+\frac{L_F}{2}\|\theta_{t+1}-\theta_t\|^2\\
	&=F(\theta_t)-\eta\langle\nabla F(\theta_t),G(\theta_t)\rangle+\frac{L_F}{2}\eta^2\|G(\theta_t)\|^2\\
	&=F(\theta_t)-\frac{1}{L_F}\langle\nabla F(\theta_t),G(\theta_t)-\nabla F(\theta_t)+\nabla F(\theta_t)\rangle\\
	&\quad+\frac{1}{2L_F}\|G(\theta_t)-\nabla F(\theta_t)+\nabla F(\theta_t)\|^2\\	
	&=F(\theta_t)-\frac{1}{2L_F}\|\nabla F(\theta_t)\|^2+\frac{1}{2L_F}\|G(\theta_t)-\nabla F(\theta_t)\|^2
	\end{split}
	\end{equation}
	where the first equality follows from $\theta_{t+1}=\theta_t-\eta\cdot G(\theta_t)$ and the second equality follows from $\eta=\frac{1}{L_F}$. Note that the derivation of (\ref*{D_1}) is a common trick in analyzing the convergence of smooth functions which shifts the burden of proving convergence into the relatively easy task of quantifying $\|G(\theta_t)-\nabla F(\theta_t)\|$.
	
	According to Lemma 2, we have $\|G(\theta_t)-\nabla F(\theta_t)\|\le C_\alpha\|\nabla F(\theta_t)\|+\Delta$ with $C_\alpha=\frac{2\alpha}{1-\beta}$ and $\Delta=L_{\theta z}\varepsilon+\sigma$, which leads to
	\begin{equation}
	\label{D_2}
	\begin{split}
	\|G(\theta_t)&-\nabla F(\theta_t)\|^2\\
	&\le C^2_\alpha\|\nabla F(\theta_t)\|^2+2C_\alpha\|\nabla F(\theta_t)\|\Delta+\Delta^2\\
	&\le (1+r)C^2_\alpha\|\nabla F(\theta_t)\|^2+(1+1/r)\Delta^2
	\end{split}
	\end{equation}
	for any $r>0$.
	
	Combining (\ref*{D_1}) and (\ref*{D_2}), we have
	\begin{equation}
	\label{D_3}
	F(\theta_{t+1})\le F(\theta_t)-\frac{1-(1+r)C^2_\alpha}{2L_F}\|\nabla F(\theta_t)\|^2+\frac{1+1/r}{2L_F}\Delta^2
	\end{equation}
	which is equivalent to
	\begin{equation}\label{D_4}
	\begin{split}
	\|\nabla F(\theta_t)\|^2\le&\frac{2L_F}{1-(1+r)C^2_\alpha}\big[F(\theta_t)-F(\theta_{t+1})\big]\\
	&+\frac{1+1/r}{1-(1+r)C^2_\alpha}\Delta^2.
	\end{split}	
	\end{equation}
	Summing up (\ref*{D_4}) for $t=0,1,\ldots,T-1$ before being divided by $T$ gives
	\begin{equation}\label{D_5}
	\begin{split}
	&\frac{1}{T}\sum_{t=0}^{T-1}\|\nabla F(\theta_t)\|^2\\
	&\le\frac{2L_F}{\big(1-(1+r)C_\alpha^2\big)T}\big[F(\theta_0)-F(\theta_T)\big]+\frac{1+1/r}{1-(1+r)C^2_\alpha}\Delta^2\\
	&\le\frac{2L_F}{\big(1-(1+r)C_\alpha^2\big)T}\big[F(\theta_0)-F(\theta^*)\big]+\frac{1+1/r}{1-(1+r)C^2_\alpha}\Delta^2
	\end{split}
	\end{equation}
	which is exactly the conclusion in Theorem 2.
	
	Note that the transition from (\ref*{D_3}) to (\ref*{D_4}) only stands under the condition that $1-(1+r)C_\alpha^2>0$, which constrains $r$ to the less than $\frac{1}{C_\alpha^2}-1$. On the other hand, $r>0$, which requires $C_\alpha=\frac{2\alpha}{1-\beta}<1$, i.e., $2\alpha+\beta<1$. Since $\beta\ge\alpha$, we can conclude that (\ref*{D_5}) holds if and only if $\alpha<\frac{1}{3}$ and $0<r<\big(\frac{1-\beta}{2\alpha}\big)^2-1$.
	
	\section{Proof of Theorem 3}
	\label{pend_E}
	First, we seek to establish the convexity of $F(\theta)$. Recall that $\phi_\lambda(\theta;x)=\sup_z\{f(\theta;z)-\lambda c(z,x)\}$. For any $\theta_1,\theta_2$ and $0\le t\le1$, we have
	\begin{equation}\label{E_1}
	\begin{split}
	&\phi_\lambda(t\theta_1+(1-t)\theta_2;x)\\
	&=\sup_z\{f(t\theta_1+(1-t)\theta_2;z)-\lambda c(z,x)\}\\
	&\le\sup_z\{tf(\theta_1;z)+(1-t)f(\theta_2;z)-\lambda c(z,x)\}\\
	&=\sup_z\{t[f(\theta_1;z)-\lambda c(z,x)]+(1-t)[f(\theta_2;z)-\lambda c(z,x)]\}\\
	&\le t\sup_z\{f(\theta_1;z)-\lambda c(z,x)\}+(1-t)\sup_z\{f(\theta_2;z)-\lambda c(z,x)\}\\
	&=t\phi_\lambda(\theta_1;x)+(1-t)\phi_\lambda(\theta_2;x)
	\end{split}
	\end{equation}
	in which the first inequality follows from Assumption 5. According to (\ref*{E_1}), $\phi_\lambda(\theta;x)$ is convex w.r.t. $\theta$. As a result, $F(\theta)=\frac{1}{N}\sum_{j=1}^{N}\phi_\lambda(\theta;x_j)$ is also convex.
	
	The convexity of $F(\theta)$ suggests that
	$F(\theta^*)\ge F(\theta_t)+\langle\nabla F(\theta_t),\theta^*-\theta_t\rangle$,
	which leads to
	\begin{equation}\label{E_2}
	\begin{split}
	F(\theta_t)-F(\theta^*)
	&\le\langle\nabla F(\theta_t),\theta_t-\theta^*\rangle\\
	&\le\|\nabla F(\theta_t)\|\cdot\|\theta_t-\theta^*\|\\
	&\le D\|\nabla F(\theta_t)\|
	\end{split}
	\end{equation}
	in which $D=k\|\theta_0-\theta^*\|$. The third inequality of (\ref*{E_2}) follows from Assumption 6. As a result, we obtain (\ref*{E_3}) as a key property in the subsequent analysis.
	\begin{equation}\label{E_3}
	\|\nabla F(\theta_t)\|\ge\frac{1}{D}\big[F(\theta_t)-F(\theta^*)\big]
	\end{equation}
	Since Theorem 3 keeps all the assumptions made in Theorem 2, all the intermediate steps in the proof of Theorem 2 also apply here. In this regard, we borrow (\ref*{D_3}), i.e.,
	\begin{equation}\label{E_9}
	F(\theta_{t+1})-F(\theta_t)\le-A\|\nabla F(\theta_t)\|^2+B
	\end{equation}
	in which we define $A=\frac{1-(1+r)C^2_\alpha}{2L_F}$ and $B=\frac{(1+1/r)(L_{\theta z}\varepsilon+\sigma)^2}{2L_F}$ for convenience.
	
	Next, we consider two cases in regard to the relationship between $A\|\nabla F(\theta_t)\|^2$ and $B$.
	
	\textbf{Case 1.} Suppose that for all $0\le t\le T-1$, it holds that $B\le\frac{A}{2}\|\nabla F(\theta_t)\|^2$. In this case, we have
	\begin{equation}\label{E_5}
	F(\theta_{t+1})-F(\theta_t)\le-\frac{A}{2}\|\nabla F(\theta_t)\|^2.
	\end{equation}
	Combining (\ref*{E_3}) and (\ref*{E_5}) gives
	\begin{equation}
	\begin{split}
	\big[&F(\theta_t)-F(\theta^*)\big]^2\\
	&\le\frac{2D^2}{A}\Big(\big[F(\theta_t)-F(\theta^*)\big]-\big[F(\theta_{t+1})-F(\theta^*)\big]\Big)
	\end{split}	
	\end{equation}
	which, after divided by $\big[F(\theta_t)-F(\theta^*)\big]\big[F(\theta_{t+1})-F(\theta^*)\big]$ on both sides, leads to
	\begin{equation}\label{E_6}
	\begin{split}
	&\frac{F(\theta_t)-F(\theta^*)}{F(\theta_{t+1})-F(\theta^*)}\\
	&\le\frac{2D^2}{A}\Big(\frac{1}{F(\theta_{t+1})-F(\theta^*)}-\frac{1}{F(\theta_t)-F(\theta^*)}\Big).
	\end{split}	
	\end{equation}
	According to (\ref*{E_5}), we have $F(\theta_{t+1})\le F(\theta_t)$. Therefore, $\frac{F(\theta_t)-F(\theta^*)}{F(\theta_{t+1})-F(\theta^*)}\ge1$, and (\ref*{E_6}) can be simplified as
	\begin{equation}\label{E_7}
	\frac{1}{F(\theta_{t+1})-F(\theta^*)}-\frac{1}{F(\theta_t)-F(\theta^*)}\ge\frac{A}{2D^2}.
	\end{equation}
	Summing up (\ref*{E_7}) for $t=0,1,\ldots,T-1$ gives
	\begin{equation}
	\begin{split}
	\frac{1}{F(\theta_T)-F(\theta^*)}&\ge\frac{AT}{2D^2}+\frac{1}{F(\theta_0)-F(\theta^*)}\\
	&\ge\frac{AT}{2D^2}
	\end{split}
	\end{equation}
	which leads to
	\begin{equation}\label{E_17}
	F(\theta_T)-F(\theta^*)\le\frac{2D^2}{AT}.
	\end{equation}
	
	\textbf{Case 2.} Suppose that there exists  $t_0\in\{0,1,\ldots,T-1\}$, such that $B>\frac{A}{2}\|\nabla F(\theta_{t_0})\|^2$. In this case, we have
	\begin{equation}\label{E_8}
	\|\nabla F(\theta_{t_0})\|<\sqrt{\frac{2B}{A}}.
	\end{equation}
	Combining (\ref*{E_3}) and (\ref*{E_8}) gives
	\begin{equation}\label{E_15}
	F(\theta_{t_0})-F(\theta^*)< D\sqrt{\frac{2B}{A}}.
	\end{equation}
	Next we show by contradiction that for all $t\ge t_0$, it holds that
	\begin{equation}\label{E_16}
	F(\theta_{t})-F(\theta^*)\le D\sqrt{\frac{2B}{A}}+B.
	\end{equation}
	Suppose that there exists $t_1\ge t_0$ such that
	\begin{equation}\label{E_10}
	F(\theta_{t_1})-F(\theta^*)> D\sqrt{\frac{2B}{A}}+B.
	\end{equation}
	According to (\ref*{E_9}), we have
	\begin{equation}\label{E_11}
	\begin{split}
	F(\theta_{t_1})-F(\theta_{t_1-1})&\le-A\|\nabla F(\theta_{t_1-1})\|^2+B\\
	&\le B.
	\end{split}
	\end{equation}
	Combining (\ref*{E_10}) and (\ref*{E_11}) gives
	\begin{equation}\label{E_12}
	F(\theta_{t_1-1})-F(\theta^*)> D\sqrt{\frac{2B}{A}}.
	\end{equation}
	Combining (\ref*{E_12}) and (\ref*{E_3}) gives
	\begin{equation}\label{E_13}
	\|\nabla F(\theta_{t_1-1})\|>\sqrt{\frac{2B}{A}}.
	\end{equation}
	Plugging (\ref*{E_13}) into (\ref*{E_9}), we obtain $F(\theta_{t_1-1})\ge F(\theta_{t_1})+B$, which suggests that (\ref*{E_10}) also holds with $t_1$ replaced by $t_1-1$. By the same token, we can conclude that (\ref*{E_10}) should hold with $t_1$ replaced by all $t\le t_1$. This is in clear contradiction with the incident of $t=t_0$ as shown in (\ref*{E_15}). Therefore, (\ref*{E_16}) is valid for all $t\ge t_0$ as stated.
	
	Finally, combining the results of Case 1 (\ref*{E_17}) and Case 2 (\ref*{E_16}), we achieve that
	\begin{equation}
	F(\theta_T)-F(\theta^*)\le\max\left\{
	\frac{2D^2}{AT},D\sqrt{\frac{2B}{A}}+B
	\right\}
	\end{equation}
	which completes the proof of Theorem 3.
	
	\section{Proof of Theorem 4}
	\label{pend_F}
	According to Lemma 1, $F(\theta)$ is $L_F$-smooth, and according to Assumption 7, $F(\theta)$ is $\lambda_F$-strongly convex. In convex optimization theory, it is well known that smooth and strongly convex functions enjoy linear convergence rate with gradient descent. Here we will first establish and then use such a property with a specific convergence factor. We start with the following equality
	\begin{equation}\label{F_1}
	\begin{split}
	&\|\theta_t-\eta\nabla F(\theta_t)-\theta^*\|^2\\
	=&\|\theta_t-\theta^*\|^2-2\eta\langle\nabla F(\theta_t),\theta_t-\theta^*\rangle+\eta^2\|\nabla F(\theta_t)\|^2.
	\end{split}	
	\end{equation}
	According to the co-coercivity of smooth and strongly convex function, we have
	\begin{equation}\label{F_2}
	\begin{split}
	\langle\nabla F(\theta_t)-\nabla F(\theta^*),\theta_t-\theta^*\rangle\ge&\frac{1}{L_F+\lambda_F}\|\nabla F(\theta_t)-\nabla F(\theta^*)\|^2\\
	&+\frac{L_F\lambda_F}{L_F+\lambda_F}\|\theta_t-\theta^*\|^2.
	\end{split}
	\end{equation}
	Since $\theta^*$ is the global minimizer of $F(\theta)$ and therefore $\nabla F(\theta^*)=0$, (\ref*{F_2}) reduces to
	\begin{equation}\label{F_3}
	\langle\nabla F(\theta_t),\theta_t-\theta^*\rangle\ge\frac{1}{L_F+\lambda_F}\|\nabla F(\theta_t)\|^2+\frac{L_F\lambda_F}{L_F+\lambda_F}\|\theta_t-\theta^*\|^2.
	\end{equation}
	Plugging (\ref*{F_3}) into (\ref*{F_1}), we have
	\begin{equation}\label{F_4}
	\begin{split}
	\|\theta_t-\eta\nabla F(\theta_t)-\theta^*\|^2\le&\left(1-2\eta\frac{L_F\lambda_F}{L_F+\lambda_F}\right)\|\theta_t-\theta^*\|^2\\
	&+\left(\eta^2-\frac{2\eta}{L_F+\lambda_F}\right)\|\nabla F(\theta_t)\|^2.
	\end{split}	
	\end{equation}
	In order to eliminate the last term in (\ref*{F_4}), we take $\eta=\frac{2}{L_F+\lambda_F}$ and simplify (\ref*{F_4}) as
	\begin{equation}
	\|\theta_t-\eta\nabla F(\theta_t)-\theta^*\|^2\le\left(1-\frac{4L_F\lambda_F}{(L_F+\lambda_F)^2}\right)\|\theta_t-\theta^*\|^2
	\end{equation}
	which is the same as
	\begin{equation}\label{F_5}
	\|\theta_t-\eta\nabla F(\theta_t)-\theta^*\|\le\frac{L_F-\lambda_F}{L_F+\lambda_F}\|\theta_t-\theta^*\|
	\end{equation}
	which verifies linear convergence with a factor of $\frac{L_F-\lambda_F}{L_F+\lambda_F}$. Note that (\ref*{F_5}) holds on condition that $\eta=\frac{2}{L_F+\lambda_F}$.
	
	Next, we try to evaluate the single-step progress made by our algorithm as follows.
	\begin{equation}\label{F_7}
	\begin{split}
	&\|\theta_{t+1}-\theta^*\|\\
	=&\|\theta_t-\eta G(\theta_t)-\theta^*\|\\
	=&\|\theta_t-\eta\nabla F(\theta_t)-\theta^*+\eta[\nabla F(\theta_t)-G(\theta_t)]\|\\
	\le&\|\theta_t-\eta\nabla F(\theta_t)-\theta^*\|+\eta\|\nabla F(\theta_t)-G(\theta_t)\|\\
	\le&\frac{L_F-\lambda_F}{L_F+\lambda_F}\|\theta_t-\theta^*\|+\frac{2}{L_F+\lambda_F}\|G(\theta_t)-\nabla F(\theta_t)\|\\
	\le&\frac{L_F-\lambda_F}{L_F+\lambda_F}\|\theta_t-\theta^*\|+\frac{2C_\alpha}{L_F+\lambda_F}\|\nabla F(\theta_t)\|+\frac{2\Delta}{L_F+\lambda_F}
	\end{split}
	\end{equation}
	in which the second inequality follows from (\ref*{F_5}) by taking $\eta=\frac{2}{L_F+\lambda_F}$, and the third inequality follows from Lemma 2 with $C_\alpha=\frac{2\alpha}{1-\beta}$ and $\Delta=L_{\theta z}\varepsilon+\sigma$.
	
	According to the properties of $F(\theta)$ being $L_F$-smooth, we have $$\frac{1}{2L_F}\|\nabla F(\theta_t)\|^2\le F(\theta_t)-F(\theta^*)\le\frac{L_F}{2}\|\theta_t-\theta^*\|^2$$ which leads to
	\begin{equation}\label{F_6}
	\|\nabla F(\theta_t)\|\le L_F\|\theta_t-\theta^*\|.
	\end{equation}
	Plugging (\ref*{F_6}) into (\ref*{F_7}), we have
	\begin{equation}\label{F_8}
	\|\theta_{t+1}-\theta^*\|\le\frac{2L_FC_\alpha +L_F-\lambda_F}{L_F+\lambda_F}\|\theta_t-\theta^*\|+\frac{2\Delta}{L_F+\lambda_F}.
	\end{equation}
	By iterating (\ref*{F_8}) we obtain
	\begin{equation}\label{F_9}
	\|\theta_{T}-\theta^*\|\le\left(\frac{2L_FC_\alpha+L_F-\lambda_F}{L_F+\lambda_F}\right)^T\|\theta_0-\theta^*\|+\frac{\Delta}{\lambda_F-L_FC_\alpha}
	\end{equation}
	which is exactly the conclusion in Theorem 4.
	
	Note that the transition from (\ref*{F_8}) to (\ref*{F_9}) only stands under the condition that $\frac{2L_FC_\alpha+L_F-\lambda_F}{L_F+\lambda_F}<1$, which requires that $C_\alpha=\frac{2\alpha}{1-\beta}<\frac{\lambda_F}{L_F}$, i.e., $2\alpha\frac{L_F}{\lambda_F}+\beta<1$. Since $\beta\ge\alpha$, we can conclude that (\ref*{F_9}) holds if and only if $\alpha<\frac{1}{1+2L_F/\lambda_F}$.

\end{document}